\RequirePackage{lineno}
\documentclass[aps,prb,preprint,groupedaddress,showkeys]{revtex4-1}
\usepackage[pdftex]{graphicx}
\usepackage[colorlinks=true,linkcolor=blue,citecolor=blue]{hyperref}
\usepackage[mathscr]{eucal}
\usepackage{amsmath,amstext,amssymb,latexsym,booktabs,amsfonts,array,bookmark,multirow}
\graphicspath{{figures/}}
\DeclareGraphicsExtensions{.pdf,.jpeg,.png}

\usepackage{url}

\newtheorem{theorem}{Theorem}[section]
                    {$\blacksquare$\vspace*{7pt}} 

\def\f{\frac}

\def\q{\quad}

\def\bi{{\mathbf i}}

\def\bi{\begin{itemize}} \def\ei{\end{itemize}}
\def\be{\begin{eqnarray*}}
\def\ee{\end{eqnarray*}}

\def\0{{\mathbf 0}}

\newcommand{\beq}{\begin{equation}}
\newcommand{\eeq}{\end{equation}}

\def\eref#1{(\ref{#1})}

\newcommand{\eps}{\varepsilon}

\def\XXint#1#2#3{{\setbox0=\hbox{$#1{#2#3}{\int}$ }
\vcenter{\hbox{$#2#3$ }}\kern-.55\wd0}}

\DeclareMathOperator*{\argmin}{arg\,min}

\begin{document}

\def\papertitle{Unpaired image denoising using a generative adversarial network in X-ray CT}
\title{\papertitle}

\author{Hyoung Suk Park}
\affiliation{National Institute for Mathematical Sciences, Daejeon, 34047, Korea.}

\author{Jineon Baek}
\affiliation{National Institute for Mathematical Sciences, Daejeon, 34047, Korea.}

\author{Sun Kyoung You}
\affiliation{Department of Radiology, Chungnam National University College of Medicine and Chungnam National University Hospital, Daejeon, 35015, Korea}

\author{Jae Kyu Choi}
\affiliation{School of Mathematical Sciences, Tongji University, China}

\author{Jin Keun Seo}
\affiliation{Department of Computational Science and Engineering,Yonsei University, Korea}

\thanks{Manuscript received XXX; revised  XXX.
Corresponding author: Jin Keun Seo (email: seoj@yonsei.ac.kr).}

\begin{abstract}	
This paper proposes a deep learning-based denoising method for noisy low-dose computerized tomography (CT) images in the absence of paired training data. The proposed method uses a fidelity-embedded generative adversarial network (GAN) to learn a denoising function from unpaired training data of low-dose CT (LDCT) and standard-dose CT (SDCT) images, where the denoising function is the optimal generator in the GAN framework. This paper analyzes  the f-GAN objective to derive a suitable generator that is optimized by minimizing a weighted sum of two losses: the Kullback-Leibler divergence between an SDCT data distribution and a generated distribution, and the $\ell_2$ loss between the LDCT image and the corresponding generated images (or denoised image). The computed generator reflects the prior belief about SDCT data distribution through training. We observed that the proposed method allows the preservation of fine anomalous features while eliminating noise. The experimental results show that the proposed deep-learning method with unpaired datasets performs comparably to a method using paired datasets. A clinical experiment was also performed to show the validity of the proposed method for noise arising in the low-dose X-ray CT.
\end{abstract}

\maketitle


\section{Introduction}

In computed tomography (CT), reducing the X-ray radiation dose,  while maintaining the diagnostic image quality is an important ongoing issue. This is because of growing concerns regarding the risk of radiation induced cancer \cite{Gonzalez2004,Miglioretti2013}. Low dose CT (LDCT) is commonly achieved by reducing the X-ray tube current. However, LDCT images obtained from commercial CT scanners in general suffer from the low signal-to-noise ratio (SNR) and the reduced diagnostic reliability. Therefore, numerous efforts have sought to denoise LDCT images, by finding a denoising mapping that converts an LDCT image to the corresponding standard-dose CT (SDCT) image.

Various iterative reconstruction (IR) methods have been proposed to reduce noise in LDCT images while preserving structure. The noise reduction modeling in these methods can employ loss functions in image space or in sinogram space. Some have incorporated prior knowledge into the denoising process by employing regularization strategies  such as total variation (TV) \cite{Dong2014,Xu2011}, fractional-order TV \cite{Zhang2016}, and nonlocal TV \cite{Liu2016}. Markov random fields theory \cite{Li2004,Wang2006} or nonlocal means \cite{Chen2009,Ma2012,Zhang2017} have also been used as prior information. Statistical image reconstruction methods such as the maximum a posteriori (MAP) approach for data fitting in sinogram space have been used for efficient noise filtering of the sinogram in LDCT \cite{Sauer1993,Li2004}.

While these IR methods can significantly improve the quality of reconstructed CT images, they retain some limitations in clinical practice. To begin with, it is challenging to design a prior that conveys the characteristics of LDCT and SDCT images. For example, commonly used priors such as TV and its variants produce an over-smoothing effect that causes the loss of fine detail such as small anomalies. Next, IR methods impose  high computational costs, as they require an iterative solver to find a reasonable approximate solution. Finally, sinogram-based methods use  projection data for data fidelity, but it is generally difficult to access projection data from  a commercial CT scanner.

Various supervised learning approaches have recently been suggested to reduce noise in LDCT images \cite{Chen2017a,Chen2017,Kang2016}. Paired training data (i.e., LDCT and SDCT images) are not available in clinical practice, because it is in general difficult to obtain both types of images simultaneously from a given patient. Therefore, the learning approaches obtain paired training data by generating simulated LDCT image data from clinical SDCT images. Results with supervised datasets have shown that these approaches can reduce noise and artifacts in the simulated LDCT images. However, their practical performance depends heavily on the quality of the simulated LDCT image data.
On the other hand,  it is easy to collect a sufficient number of "unpaired" SDCT and LDCT images from an in-hospital database, as the use of LDCT increases in routine clinical practice \cite{Chang2013}.

Motivated by the absence of paired data in the clinical environment, this study uses an unpaired learning approach to develop a noise reduction function that maps from LDCT images to its SDCT-like image counterparts. We exploit a method of MAP estimation for LDCT image denoising that imposes a compromise between noise reduction and a model-fitness with SDCT-like image priors obtained from the data of SDCT images. The problem of the MAP estimation can be converted into a minimization problem involving cross-entropy of data distribution and the distribution generated by the generative adversarial network (GAN) \cite{Goodfellow2014}. Given the general difficulty of directly handling cross-entropy, it is approximated to the Kullback-Leibler(KL) divergence, thus allowing an approximate MAP estimation using the GAN framework. This facilitates training of the network architecture with unpaired datasets.

\begin{figure*}[ht!]
	\centering
	\includegraphics[width=0.8\textwidth]{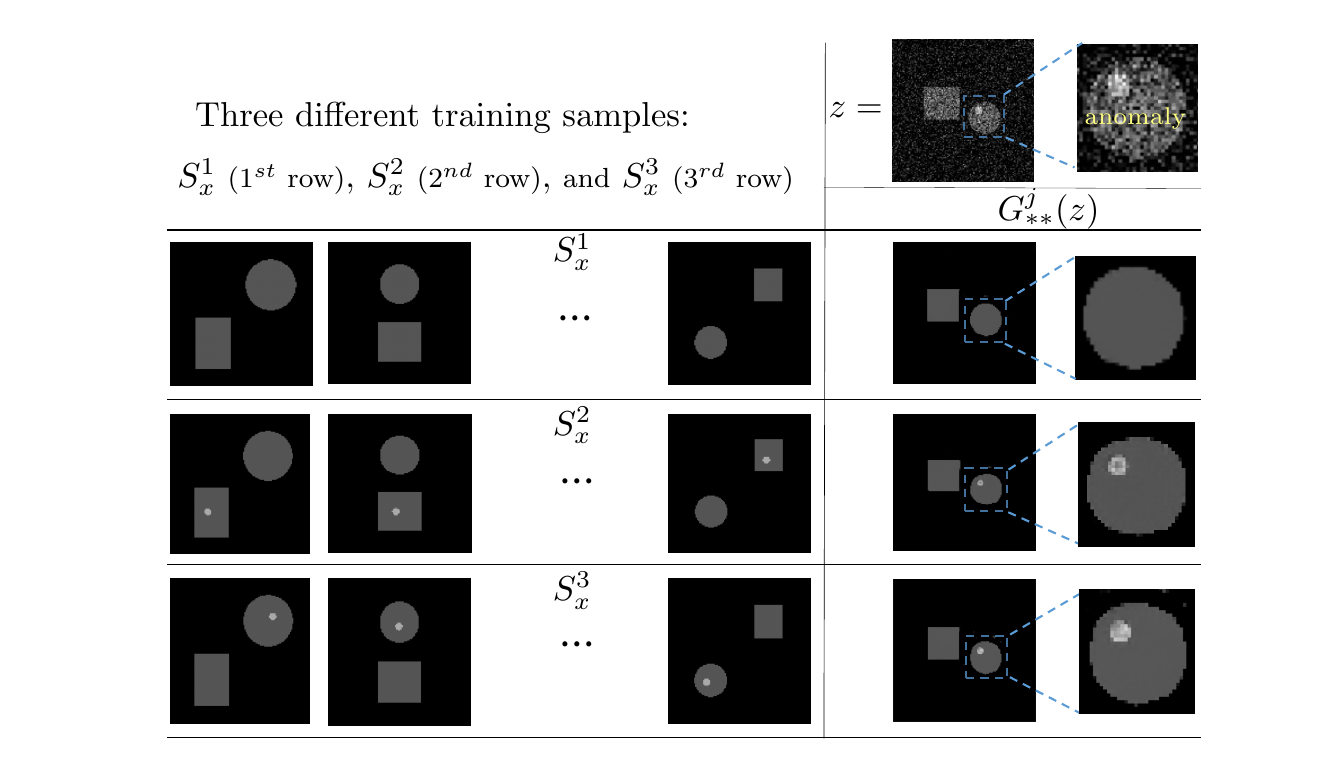}
	\caption{Learning ability of the proposed method. Given $z$, which contains a small anomaly in the disk (top right), three different training samples $\mathcal{S}_{x}^{1},~\mathcal{S}_{x}^{2}$, and $\mathcal{S}_{x}^{3}$  are used to show the data-driven dependency.  The generator $G_{**}^1(z)$ eliminates the anomaly, because there is no small anomaly in the training samples $\mathcal{S}_{x}^{1}$. On the other hand,  both $G_{**}^2(z)$ and $G_{**}^3(z)$  preserve the small anomaly in the disk well while removing the noise. It is quite interesting to see that $\mathcal{S}_{x}^{2}$ contains the anomaly only inside the rectangle (not inside the disk).}
	\label{map-prior}
\end{figure*}

The proposed GAN approach is capable of inferring the desired SDCT-like image priors from the data of the SDCT images. It is critical to choose effective training datasets to reflect the appropriate image priors, in order to preserve detailed features of the image while eliminating noise. This data dependency of SDCT image priors is validated by the experiment shown in Fig.   \ref{map-prior}. Under the assumption of Gaussian noise, the $\ell_2$ constraint is added to the proposed GAN model. Unlike conventional approaches that treat the entire image, the proposed method is carried out patch-by-patch manner, which can effectively train local noise features in the CT images. Numerical simulation and clinical results demonstrate that the proposed method has a great potential to reduce noise in LDCT images for unpaired CT images or even for a complicated noise model.

\section{Method}\label{Method}
Throughout this paper, we denote by $z$ and $x$ LDCT and SDCT images, respectively, both of which are $n\times n$ pixel grayscale images.  We assume that unpaired training samples $\{ z_k\}_{k=1}^N$ and $\{ x_k\}_{k=1}^M$ are drawn from unknown LDCT data distribution $p_z(z)$  and unknown SDCT data distribution $p_x(x)$, respectively. Here, $N$ and $M$ denote the number of LDCT and SDCT training samples, respectively. In terms of image denoising, $z \sim p_z$ corresponds to  noisy image (source) and $ x \sim p_x$ is regarded as noise-free image (target).

The object of our method for denoising is to learn a generator $G$ with input $z \sim p_z$ 
using GAN's framework together with the unpaired training data, in such a way that the generator's distribution $p_{G(z)}$ is an optimal approximation of the clean image distribution $p_x$ and the generators's fidelity $\| z-G(z)\|_2$ is reasonably small for every $z\sim p_z$. Here, $\|z\|_2$ stands for the standard $\ell_2$-norm of $z$.

We adopt the additive Gaussian noise model that the noisy LDCT image $z$ is decomposed into a desired denoised image $x^*\sim p_x$ and additive Gaussian noise $\eps$ of variance $\sigma$ at each pixel, i.e., $z=x^*+\eps$ with $\eps\sim \mathcal N(0,\sigma)$.  One can estimate the clean image $x^*$ in terms of maximum a posteriori approach: Given $z \sim p_z$, find $\hat x$ that maximizes the conditional probability $p_{x^*|z}(\hat{x}|z)$ \cite{Sonderby2017}. This $\hat x$ is $\arg\max_{y} L_z(y)$, where
\begin{align}\label{map_gaussian}
 L_z(y):=\log p_{x}(y) -\lambda \|y-z\|_2^2
\end{align}
and $\lambda=\f{1}{2\sigma^2}$. For more detail, we refer to Appendix \eref{map-derivation}. However, this approach cannot be used directly due to the unknown distribution $p_{x}$. Instead of solving the denoising problem one-by-one for each $z\sim p_z$, we look for the denoising function $G: z\to \hat x$ by using GAN's framework together with many samples $\{ z_k\}_{k=1}^N$ and $\{ x_k\}_{k=1}^M$.

An optimal denoising function can be derived by maximizing the expectation $E_{z \sim p_z} L_z(G(z))$  with respect to generator $G$:
\begin{align}\label{map_gaussian2}
G_{\mbox{\tiny MAP}} :=\arg\min_G E_{z \sim p_z}\left[ - \log(p_x(G(z))) + \lambda ||G(z) - z||^2\right].
\end{align}
With the notion of  the cross entropy $H(p, q)$, we  have the following expression:
\begin{align}\label{entropy}
G_{\mbox{\tiny MAP}} &= \arg\min_G  - E_{G(z) \sim p_{G(z)}} \left[\log(p_x(G(z)))\right]\nonumber\\
&\hspace{1.6cm}+ \lambda E_{z \sim p_z}\left[ ||G(z) - z||^2\right] \nonumber\\
& = \arg\min_G H(p_{G(z)}, p_x) +  \lambda E_{z \sim p_z}\left[ ||G(z) - z||^2\right]
\end{align}
However, it is still difficult to handle the cross-entropy $H(p_{G(z)}, p_x)$ only from the training samples even if we have a complete information of $G$. Instead, we approximate  $G_{\mbox{\tiny MAP}}$  by
\begin{align}\label{kl-divergence}
G_* :=\arg\min_G KL(p_{G(z)}, p_x) + \lambda E_{z \sim p_z} \left[||G(z) - z||^2\right]
\end{align}
where the cross-entropy $H(p_{G(z)}, p_x)$ in \eqref{entropy} is replaced by KL-divergence $KL(p_{G(z)}, p_x) = H(p_{G(z)}, p_x) - H(p_{G(z)})$. Interested readers may refer to the paper\cite{Sonderby2017} for the effect of this replacement.

Now, we are ready to explain our fidelity-embedded GAN model. In GAN's framework, the unpaired datasets are used  to learn the generator $G$ that minimizes $KL(p_{G(z)}, p_x)$, while the discriminator $D$ tries to distinguish between generated $G(z)$ and $x\sim p_x$. The following theorem presents the proposed model which solves the optimal  $G_*$ in \eref{kl-divergence} exactly when $G$ and $D$ reach their optimum.
\begin{theorem}\label{tm-main}
  The optimal $G_*$ in \eref{kl-divergence} is achieved by
\begin{align}\label{our_model0}
G_*= \argmin_G \max_D  J(D,G)
\end{align}
where
\begin{align}\label{our_model1}
 J(D,G):= E_{x \sim p_x} \left[D(x)\right] &+ E_{z \sim p_z}\left[\log(1-D(G(z)))\right]\nonumber\\
 &+ \lambda E_{z \sim p_z}\left[||G(z) - z||^2\right]
\end{align}
\end{theorem}
\textit{Proof}. ~  Given a fixed $G$, we first maximize $J(D,G)$ in \eref{our_model1} with respect to $D$.
Since the third term in \eref{our_model1} is independent of $D$, it is enough to find
\begin{align}\label{our_model1-2}
D_{opt} = \arg\max_D  E_{x \sim p_x} \left[D(x)\right] + E_{z \sim p_z}\left[\log(1-D(G(z)))\right]
\end{align}
Writing $x$ and $G(z)$ in \eref{our_model1-2} to the variable $w$, we have
\begin{align}
 D_{opt} & = \arg\max_D  E_{w \sim p_x} \left[D(w)\right] + E_{w \sim p_{G(z)}}\left[\log(1-D(w))\right] \nonumber \\
 & = \arg\max_D \int p_x(w) D(w) + p_{G(z)}(w) \log(1-D(w)) dw \label{our_model:integrand}
\end{align}
where the integral spans over the space of all images.

For a fixed $w$, the value of $D(w) = D_{opt}(w)$ that maximizes the integrand can be shown to be
\begin{align}
D_{opt}(w) := 1 - \f{p_{G(z)}(w)}{p_x(w)}, \label{our_model:result}
\end{align}
because $t=D_{opt}(w)$ satisfies
\begin{align}
\frac{d}{d t} \left(p_x(w) t + p_{G(z)}(w) \log(1-t)\right)= 0.
\end{align}
Then, it follows from \eref{our_model1} and \eref{our_model:result} that
\begin{align}\label{our_model1-3}
& E_{x \sim p_x} \left[D_{opt}(x)\right] + E_{z \sim p_z}\left[\log(1-D_{opt}(G(z)))\right] \nonumber\\
&\hspace{2.4cm}+ \lambda E_{z \sim p_z}\left[||G(z) - z||^2\right] \nonumber \\
& = \int \left[ p_x(w) D_{opt}(w) + p_{G(z)}(w) \log(1-D_{opt}(w)) \right] dx \nonumber\\
&\hspace{3.15cm}+ \lambda E_{z \sim p_z} ||G(z) - z||^2 \nonumber \\
& = \int \left[ p_x(w) - p_{G(z)}(w) + p_{G(z)}(w) \log\left(\frac{p_{G(x)}(w)}{p_x(w)}\right) \right] dw\nonumber\\
&\hspace{3.7cm}+ \lambda  E_{z \sim p_z}||G(z) - z||^2 \nonumber \\
& = KL(p_{G(z)}, p_x) +  \lambda E_{z \sim p_z} ||G(z) - z||^2,
\end{align}
where $KL(p,q)$ is the KL-divergence given by $KL(p,q):=\int p(x)\log\left(\f{p(x)}{q(x)}\right)dx$. Here, the last equality follows from the fact that $\int p_x(w)dw=\int p_{G(z)}(w)dw=1$. This completes the proof.


\subsection{Data-driven dependency}
The unpaired training samples $\mathcal{S}_{z} = \{ z_k\}_{k=1}^N$ and $\mathcal{S}_{x} = \{ x_k\}_{k=1}^M$ are used to  compute the generator $G_*$ approximately:
\begin{align}\label{model-practice}
 G_{**}= &\arg\min_G \f{1}{N} \sum_{z\in \mathcal{S}_{z}}\left[\log(1-D(G(z))) + \lambda \|G(z)-z\|^2\right]
\end{align}
with $D=\arg\min_D \f{1}{M} \sum_{x\in \mathcal{S}_{x}}D(x) + \f{1}{N}$ $\sum_{z\in \mathcal{S}_{z}} \log(1-D(G(z)))$.

It should be noted that the computed generator $G_{**}$ depends heavily on the training samples. Fig.   \ref{map-prior} shows the data-driven dependency, where we use three different training samples $\mathcal{S}^j=\{\mathcal{S}_{z}, \mathcal{S}^j_{x}\}, j=1,2,3$. The first training sample $\mathcal{S}^1$ consist of one disk and one rectangle with different sizes and positions (first column).  The other samples ($\mathcal{S}^2$ and $\mathcal{S}^3$) are generated by adding a small anomaly to sample $\mathcal{S}^1$. In $\mathcal{S}^2$, the anomaly is added only in the rectangle ($\mathcal{S}^2$).  On the other hand, $\mathcal{S}^3$ is generated by adding the anomaly only in the disk. In this case,  the positions of the anomaly are randomly chosen.  Fig.   \ref{map-prior}  shows the comparison of the performance of  three generators $G_{**}^{1},~G_{**}^{2}$, and $G_{**}^{3}$, where $G_{**}^{j}$ is computed by \eref{model-practice} with $ \mathcal{S}$ replaced by $ \mathcal{S}^j$. The top rightmost image in Fig.   \ref{map-prior} is a test image $z \sim p_z$, which contains an anomaly inside the disk. The  $G_{**}^1(z)$  cannot preserve the anomaly inside the disk (the first row), whereas both $G_{**}^2(z)$ and $G_{**}^3(z)$ can preserve anomaly well. It is quite informative that $G_{**}^2(z)$ preserves the anomaly even though the training sample does not contain the anomaly inside the disk (the second row).



In this instance, the MAP estimation was completed in a patch-by patch manner, which  is clearly explained in \ref{patch-method}.

\subsection{Patch-based Network for Practical Applications}\label{patch-method}

In practice, the proposed method is executed in a patch-by-patch manner, rather than working with the entire  images (e.g., $512\times 512$ pixel CT images). Here, we take advantage of the fact  \cite{Peyre2008, Peyre2009} that the patch manifolds of many images have a low dimensional structure. Moreover, the number of training data sets is significantly increased. These allow us to learn the generated distribution efficiently over the patches extracted from SDCT images. For simplicity, we will use the same notation of $x$ and $z$ to represent the image patches (there will be no confusion from the context).

\begin{figure}[ht]
	\centering
	\includegraphics[width=0.8\textwidth]{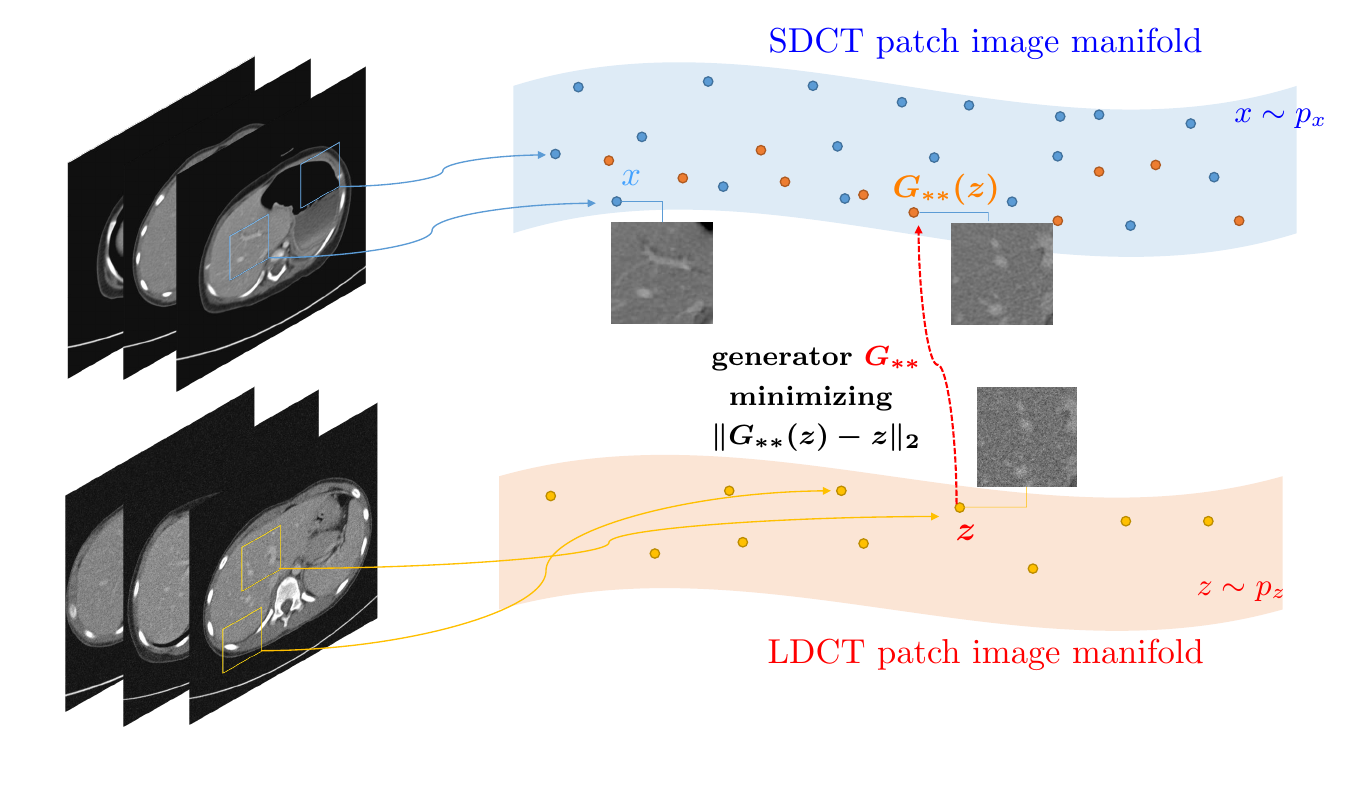}
	\caption{Schematic of generated distribution containing image priors for patches extracted from SDCT images.}
	\label{fig_ovserv}
\end{figure}

Fig.   \ref{fig_ovserv} illustrates how the proposed method in \eref{model-practice} generates $G_{**}(z)$ from the unpaired training image patches. It optimizes $p_{G_{**}(z)}$ by forcing it to be close to $p_x$ while also minimizing the $\ell_2$ distance $\|G_{**}(z)-z\|_2$. Fig.   \ref{efficiency} shows the performance of the  $G_{**}$ with two different training samples; image patches of $\mathcal{S}^1$ and $\mathcal{S}^2$  are randomly selected, respectively, from 50 different CT slices (third column) and 600 different CT slices ($x$) (fourth column), respectively.  As shown in Fig.   \ref{efficiency}, the performance of $G_{**}^{1}$ (using $\mathcal{S}^1$) is inferior to that of $G_{**}^{2}$ (using $\mathcal{S}^2$), owing to the lack of the diversity of $x$, which is called `mode collapse' \cite{Che2017,Salimans2016}.

\begin{figure*}[ht]
	\centering
	\includegraphics[width=0.8\textwidth]{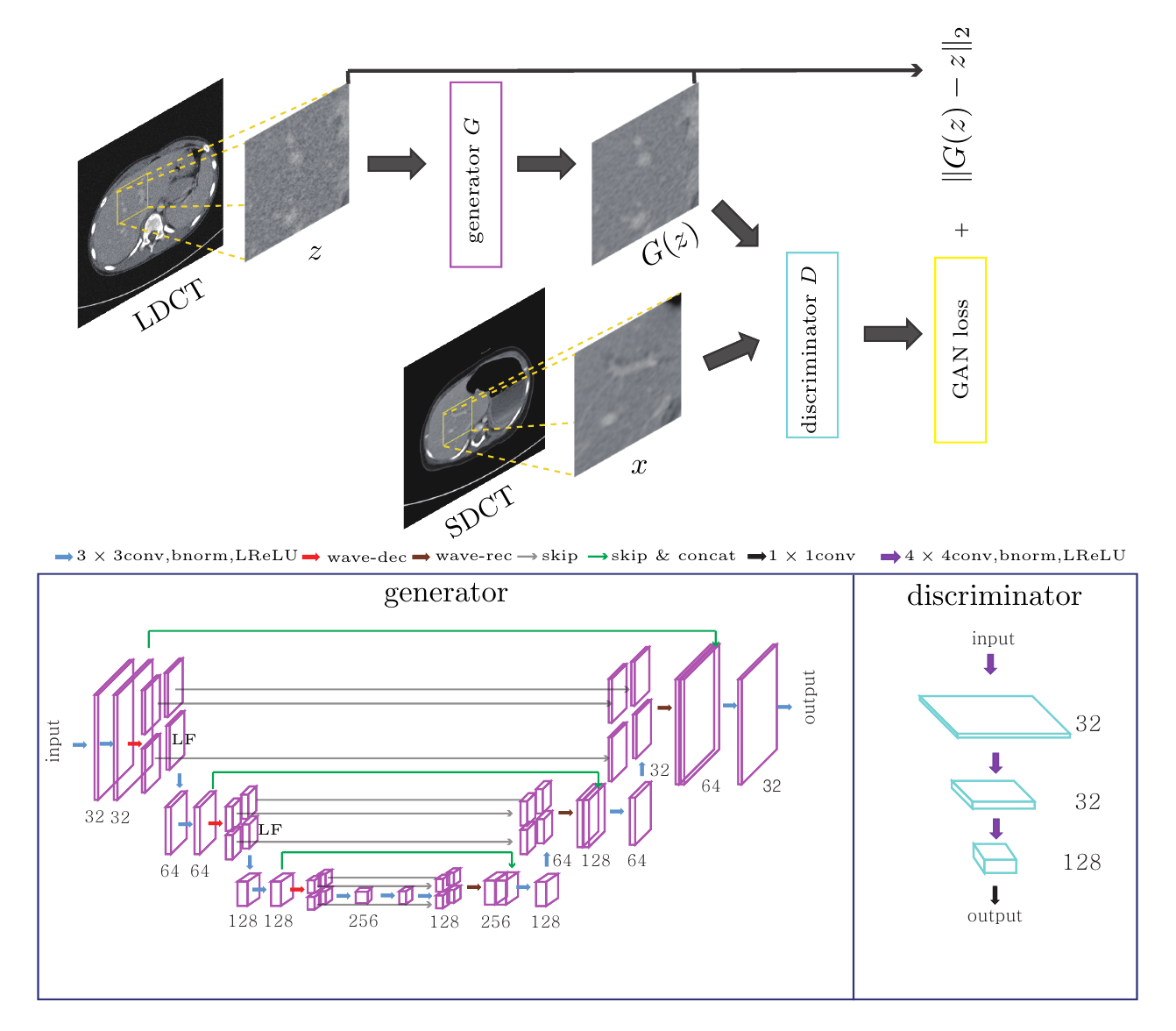}
	\caption{Network architecture of the proposed network.}
	\label{network}
\end{figure*}

\subsection{Network Architecture}
For our generator, we adopt the deep convolutional framelet \cite{Ye2017}, which is a multi-scale convolutional neural network, consisting of a contracting path and an expansive path with skipped connections and a concatenation (concat) layer. Each step of the contracting and expansive paths consists of two repeated convolutions (conv) with a $3\times3$ window.  Each convolution is followed by a batch normalization (bnorm) and a leaky rectified linear unit (LReLU). Downsampling and upsampling of the features are performed by  2-D Haar wavelet decomposition (wave-dec) and recomposition (wave-rec), respectively. High pass filters after wavelet decomposition skip directly to the expansive path, while low pass filters (marked by `LF' in Fig.   \ref{network}) are concatenated with the features in the contracting path during the same step. At the end, an additional convolution layer is added to generate a grayscale output image. Note that each convolution in our network is performed with zero-padding to match the size of the input and output images. The architecture of the deep convolutional framelet is quite similar to that of the U-net \cite{Ronneberger2015}, a deep learning model widely used in image processing, except for the high pass filter pass. The deep convolutional framelet utilizes the wavelet decomposition and recomposition instead of pooling and unpooling operations for downsampling and upsampling, respectively. By doing so, more detailed information of image can be preserved during downsampling \cite{Ye2017}.

In an adversarial architecture, we adopt the PatchGAN classifier \cite{Isola2017,Li2016,Zhu2017} as a discriminator, which tries to classify whether each patch in an image is real or fake.  PatchGAN enables us to learn the structural  detail in images more easily than does the $1 \times 1$ PixelGAN. The discriminator contains three convolution layers with a $4\times4$ window and strides of two in each direction of the domain and each layer is followed by a batch normalization and a leaky ReLU with a slope of 0.2. As the final stage of the architecture,  a $1\times 1$ convolution layer is added to generate 1-dimensional output data. Fig.   \ref{network} illustrates the architecture of the proposed method.

In our numerical simulation and clinical experiment, the objective function in \eref{model-practice} is minimized using an Adam optimizer \cite{Kingma2014} with a learning rate of 0.0002 and mini-batch size of 40, and 300 epochs are utilized for training. Training was implemented using Tensorflow \cite{Abadi2016} on a GPU (NVIDIA, Titan Xp. 12GB) system. It takes about a day to train the network. We empirically choose $\lambda=10$ in Eq. \eref{model-practice}. The network weights were initialized following a Gaussian distribution with a mean of 0 and a standard deviation of 0.01.

\begin{figure*}[ht!]
	\begin{center}\includegraphics[width=1\textwidth]{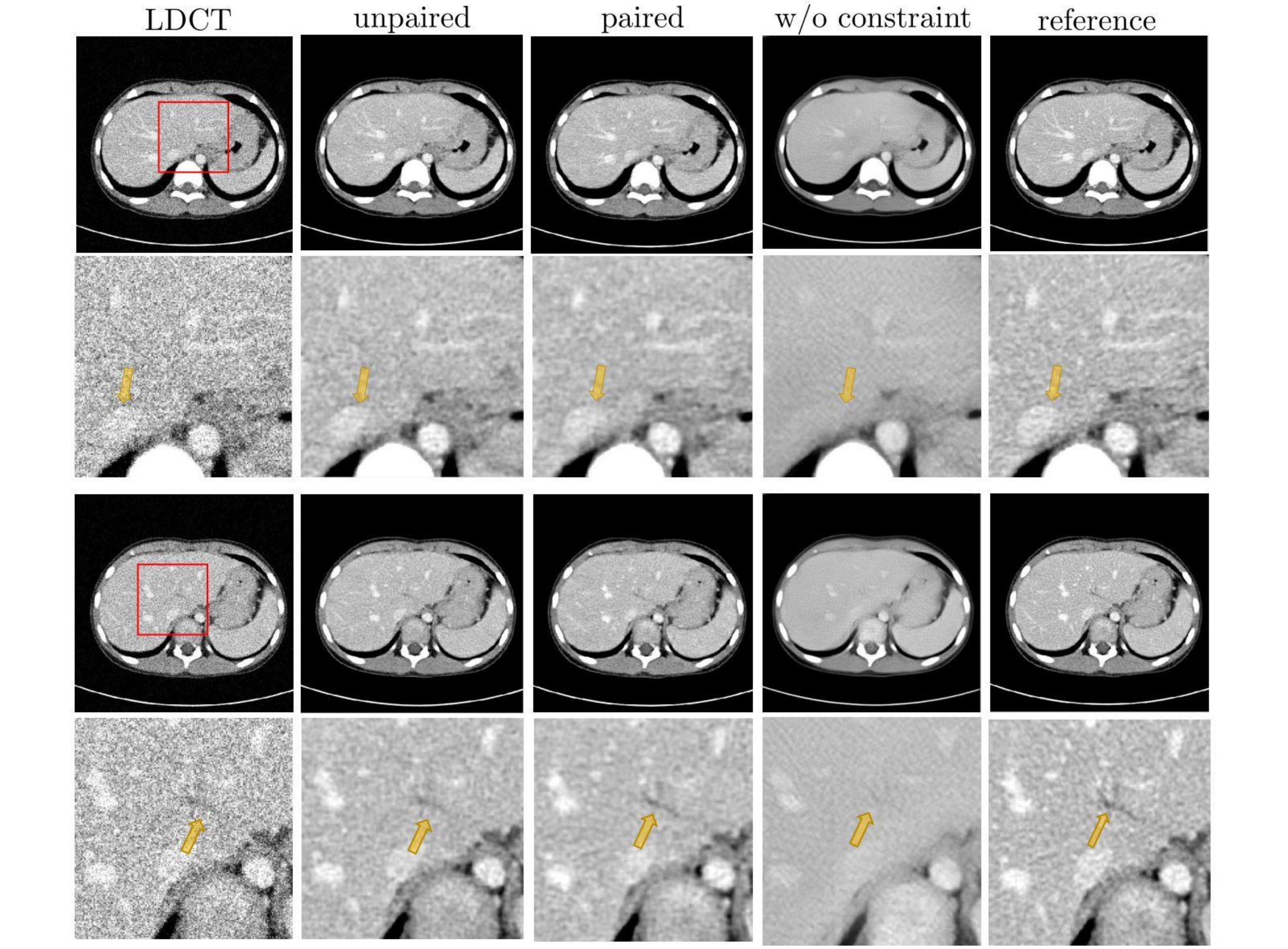}\end{center}
	\caption{Simulation results for LDCT images with Gaussian noise. The proposed unpaired learning method (second column) shows a similar performance to that of the paired learning method (third column). On the other hand, the proposed method without the fidelity term $\|G_{**}(z)-z\|_2$ removes small details as shown in the fourth column.The second and fourth rows show the zoomed ROIs of the rectangular regions of images in the first and third rows, respectively.  (C=50 HU/ W=300 HU for all CT images and zoomed ROIs)}
	\label{simu_images}
\end{figure*}

\subsection{Datasets}

In our simulations and clinical experiments, all CT images of size $512\times 512$ are acquired by a 64-channel multi-detector CT scanner (Sensation 64; Siemens Healthcare, Forchheim, Germany).

\subsubsection{Simulation study}
\begin{figure}[ht!]
	\begin{center}\includegraphics[width=0.8\textwidth]{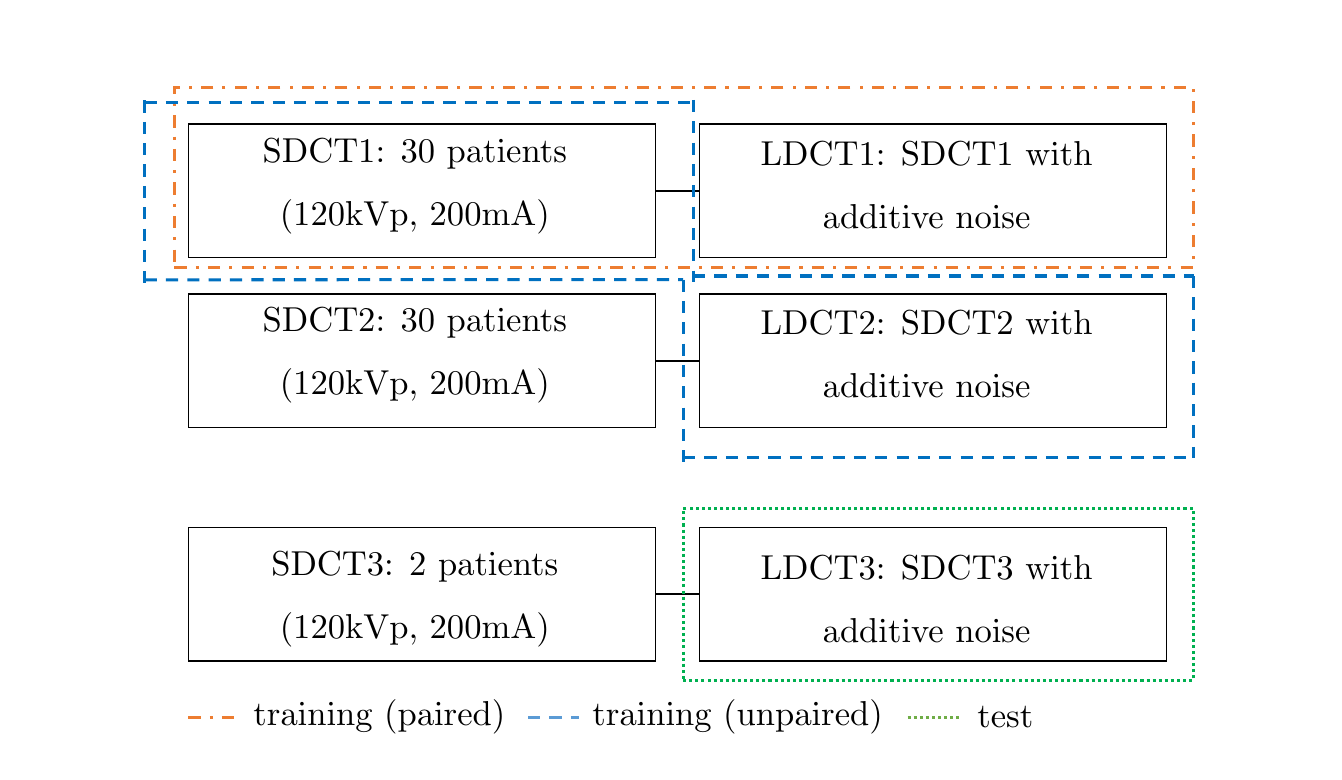}\end{center}
	\caption{Simulation datasets used for training and test.}
	\label{Simulation_datasets}
\end{figure}

We collected SDCT images of 62 patients acquired at a tube voltage of 120 kVp and a tube current-time product of 200 mAs. For each patient, we select 20 CT images that included the liver and portal vein. LDCT images are generated by adding noise to the SDCT images of the 62 patients. For the chosen SDCT images of 30 patients, corresponding LDCT images, and LDCT images of another 30 patients are used for the paired and unpaired training datasets, respectively. The LDCT images of the remaining two patients (not used for training) are used as a test dataset. See Fig.  \ref{Simulation_datasets}. From each image of size $512 \times 512$, the patches of size $128\times 128$ are extracted with strides of 8 in each direction in the image domain. Among them, 40 patches are randomly selected, and they are used as training image patches.

In this study, we generate LDCT images with two different types of noise; Gaussian noise with a standard deviation of 25 (HU), and  quantum noise based on the CT acquisition model. We generate a synthetic sinogram according to the Poisson quantum noise \cite{zeng2010}. In our simulations, the blank scan flux per detector was set to $8\times10^4$, similar to \cite{Chen2017a}. Electronic noise, modeled as Gaussian noise \cite{LaRiviere2005}, is also added to the synthetic sinogram. LDCT images are reconstructed by the filtered backprojection \cite{Bracewell1967}, where we only considered mono-chromatic and parallel X-ray beams.

\subsubsection{Clinical study}
For training our network, we collect 200 brain CT images of 10 patients, acquired at a tube voltage of 120 kVp and a fixed tube current-time product of 200 mAs. We further collect another 200 CT images at a tube voltage of 100 kVp and  a reference tube current-time product of 190 mAs. We refer to the former and latter images as low-dose CT (LDCT) images, and standard-dose CT (SDCT) images, respectively. For LDCT images (obtained with a reference mAs of 190), the mAs varies from 156 mAs to 200 mAs along the length of the scan, due to the use of an automatic exposure compensation system.  Fig.   \ref{examples_unpair} shows unpaired brain CT images used for training.  Using a low tube voltage and a tube current increases the image noise as shown in Fig.   \ref{examples_unpair}. Sixty LDCT images of three other patients who are not used in the training were collected for testing. As in the simulation, we generate unpaired patches of size $128\times 128$ from both LDCT and SDCT images.

\begin{figure*}[ht!]
	\centering
	\includegraphics[width=0.9\textwidth]{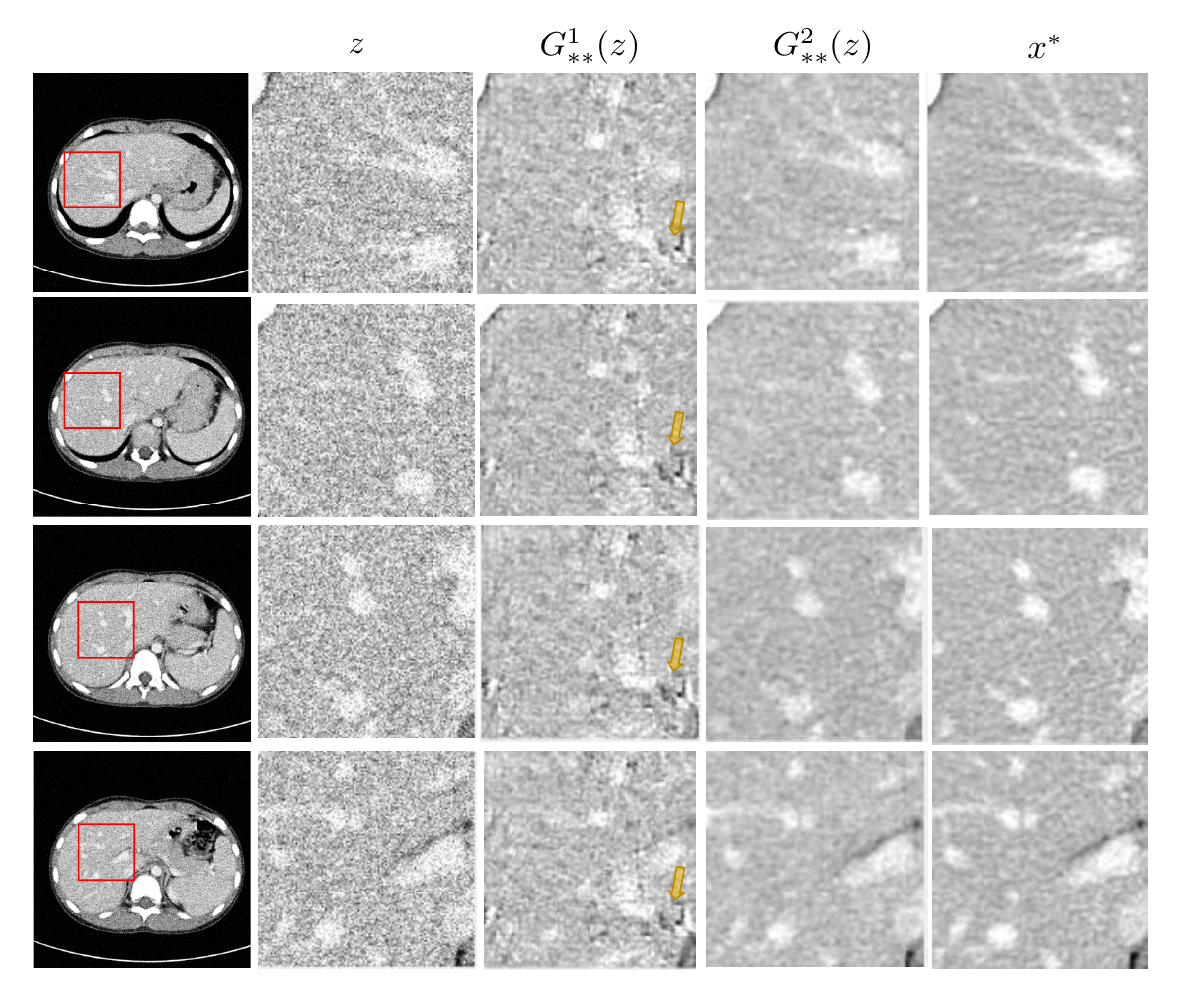}
	\caption{Results of the proposed method with two different training samples $ \mathcal{S}^1$ and $\mathcal{S}^2$ for Gaussian noise. $\mathcal{S}^1$ and $\mathcal{S}^2$ are sets of 24000 patches that are randomly selected, respectively, from 50 different SDCT slices (third column) and 600 different SDCT slices (fourth column). The performance of $G_{**}^{2}$ (using $\mathcal{S}^2$) is suprior to that of $G_{**}^{1}$ (using $\mathcal{S}^1$), due to the diversity of training samples. The third column shows the mode collapse phenomenon  \cite{Che2017,Salimans2016} in the areas that are pointed to by the yellow arrows. (C=50 HU/ W=300 HU for all CT images and patches)}
	
	\label{efficiency}
\end{figure*}

\begin{figure*}[ht!]
	\centering
	\includegraphics[width=1\textwidth]{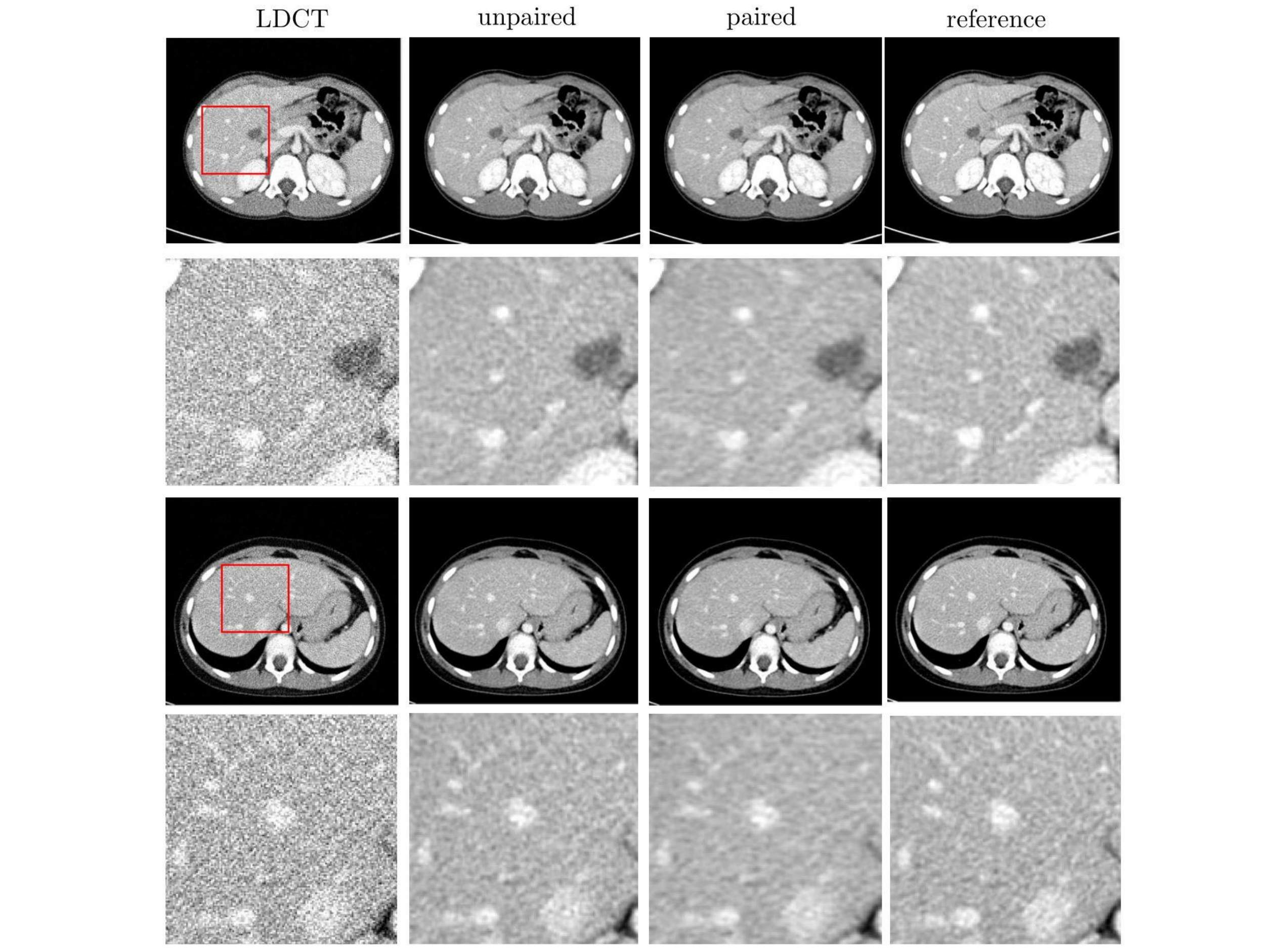}
	\caption{Simulation results for LDCT images with quantum noise. The proposed unpaired learning method (second column) shows a similar performance to that of the paired learning method (third column). The second and fourth rows show the zoomed ROIs of the rectangular regions of images in the first and third rows, respectively.  (C=50 HU/ W=300 HU for all CT images and patches)}
	\label{simu_results_quantum noise}
\end{figure*}

\begin{figure*}[ht!]
	\centering
	\includegraphics[width=0.9\textwidth]{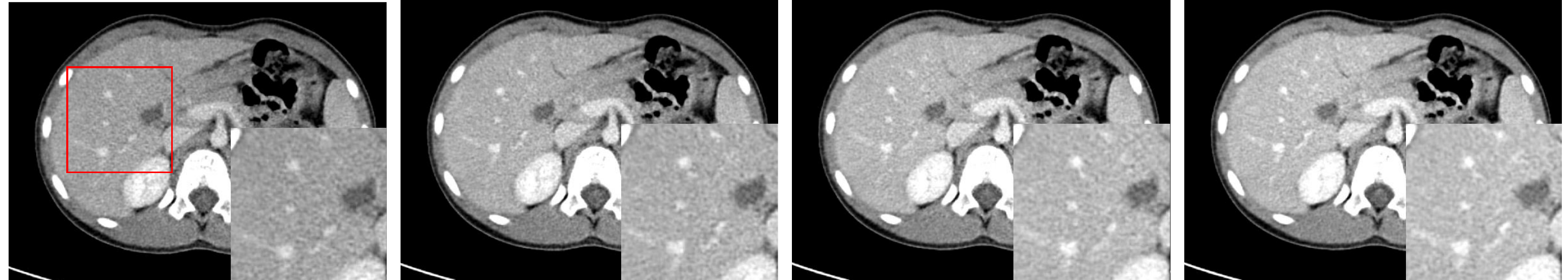}
	\caption{Performance of the proposed method with various levels of quantum noise. For training, we use LDCT images with blank scan flux of $I_0=8\times 10^4$. We use test LDCT images (in Fig. \ref{simu_results_quantum noise}) with different  $I_0= 5\times 10^4, 6.5\times 10^4, 8\times 10^4,$ and $9.5\times 10^4$ (from left to right). See Table \ref{table2}  for quantitative evaluations. (C=50 HU/ W=300 HU for all CT images and patches)}
	\label{simu_results_different noise}
\end{figure*}

\section{Results}\label{Results}

The numerical simulation and the clinical experiment are performed in order to demonstrate the validity of the proposed approach for noise reduction in CT images.

Fig.   \ref{simu_images} and Fig.   \ref{simu_results_quantum noise} show the denoising results for liver CT images with simulated Gaussian and quantum noise, respectively. The second and third columns show the Results of the proposed method using paired and unpaired LDCT and SDCT datasets, respectively. The second and fourth rows show the zoomed regions of interest (ROI) of the rectangular regions of the images in the first and third rows, respectively. As shown in Fig.   \ref{simu_images} and Fig.   \ref{simu_results_quantum noise}, both unpaired and paired learning methods reduce noise artifacts clearly, while preserving the morphological structures of the tissues. In Fig.   \ref{simu_images}, the fourth column shows the results of the model of \eref{model-practice} without the fidelity term $\|G_{**}(z)-z\|_2$. The method without the constraint generates plausible SDCT images, but tends to remove small details. Note the yellow arrow in the fourth column of Fig.   \ref{simu_images}.

Table \ref{table} shows the comparison between the results from executing the proposed method with Gaussian and quantum noise, using paired and unpaired datasets. The average peak signal to noise ratio (PSNR), structure similarity (SSIM), and mean square error (MSE) between the reference and LDCT/corrected images are calculated for the test dataset (40 LDCT images of 2 patients). The performance of the proposed method with unpaired datasets shows comparable image quality (or structure similarity) to that with paired datasets in terms of PSNR, MSE and SSIM. Overall, compared with the quality of the LDCT images, the performance of the proposed method with Gaussian noise is slightly better than that with quantum noise. This is because  fidelity in \eref{model-practice} is optimal for Gaussian noise.

\begin{table*}[ht]
\caption{PERFORMANCE COMPARISON BETWEEN THE RESULTS FROM EXECUTING THE PROPOSED METHOD WITH GAUSSIAN and QUANTUM NOISE, USING PAIRED AND UNPAIRED DATASETS.}
\centering
\begin{tabular}{|c|c|c|c|c|c|c|}
\hline
\multirow{2}{*}{Measures} & \multicolumn{3}{c|}{Guassian noise} & \multicolumn{3}{c|}{Quantum noise} \\ \cline{2-7}
                          & \quad\quad LDCT \quad\quad   & \quad\quad paired \quad\quad    & \quad\quad unpaired \quad\quad   & \quad\quad LDCT  \quad\quad    & \quad\quad paired \quad\quad    &  \quad\quad unpaired  \quad\quad \\ \hline
PSNR(dB)                  &  24.3305 &  35.2691   &    34.8404  & 23.9892   & 33.5025    & 33.4212    \\ \hline
SSIM                      &  0.2600  &  0.9154    &     0.9147  & 0.3176    & 0.8831     & 0.8688     \\ \hline
MSE                       &    1.3284e+03  &  109.4953  &   120.2068  & 1.4368e+03    & 161.0430   & 164.7475   \\ \hline
\end{tabular}
\label{table}
\end{table*}

\begin{table*}[ht]
\caption{Performance comparison of the proposed method for the test datasets acquired at different noise levels}
\centering
\begin{tabular}{|c|c|c|c|c|c|c|c|c|}
\hline
        & \multicolumn{2}{l|}{$I_0=5*10^4$} & \multicolumn{2}{l|}{$I_0=6.5*10^4$} & \multicolumn{2}{l|}{$I_0=8*10^4$} & \multicolumn{2}{l|}{$I_0=9.5*10^4$} \\ \hline
measure & LDCT             & unpair         & LDCT              & unpair          & LDCT             & unpair         & LDCT              & unpair          \\ \hline
PSNR    & 22.1787          & 27.9628        & 23.1996           & 31.3124         & 23.9892          & 33.4212        & 24.6277           & 33.8760         \\ \hline
SSIM    & 0.2425           & 0.8639         & 0.2826            & 0.8756          & 0.3176           & 0.8688         & 0.3486            & 0.8884          \\ \hline
MSE     & 2.1801e+03       & 578.6777       & 1.7233e+03        & 266.5740        & 1.4368e+03       & 164.7475       & 1.2404e+03        & 150.9356        \\ \hline
\end{tabular}
\label{table2}
\end{table*}

\begin{table*}[]
\caption{Quantitative evaluation of the proposed method for clinical brain CT images.}
\centering
\begin{tabular}{|c|c|c|c|}
\hline
 \q\q\q\q     & \begin{tabular}[c]{@{}c@{}}\q\q LDCT images\q\q\end{tabular} & \begin{tabular}[c]{@{}c@{}}\q\q Corrected images \q\q \end{tabular} & \q\q p-value  \q\q        \\ \hline
\multicolumn{4}{|l|}{CT number (Mean$\pm$SD)}                                                                                                     \\ \hline
\q\q\q\q GM \q\q\q\q   & 32.45$\pm$1.22                                            & 31.78$\pm$1.06                                                 & \textless{}0.001 \\ \hline
\q\q\q\q HM \q\q\q\q   & 26.36$\pm$1.52                                            & 26.34$\pm$1.08                                                 & 0.81             \\ \hline
\multicolumn{4}{|l|}{SNR (Mean$\pm$SD)}                                                                                                           \\ \hline
\q\q\q\q GM \q\q\q\q   & 4.13$\pm$0.48                                             & \begin{tabular}[c]{@{}c@{}}6.44$\pm$0.69\end{tabular}       & \textless{}0.001 \\ \hline
\q\q\q\q HM \q\q\q\q   & \begin{tabular}[c]{@{}c@{}}3.49 $\pm$0.41\end{tabular}  & \begin{tabular}[c]{@{}c@{}}5.53$\pm$0.59\end{tabular}       & \textless{}0.001 \\ \hline
\multicolumn{4}{|l|}{CNR (Mean$\pm$SD)}                                                                                                           \\ \hline
GM-HM & \begin{tabular}[c]{@{}c@{}}0.56 $\pm$0.19\end{tabular}  & \begin{tabular}[c]{@{}c@{}}0.78$\pm$0.23\end{tabular}       & \textless{}0.001 \\ \hline
\end{tabular}
\label{table3}
\end{table*}

We provide a performance evaluation of the proposed method for different noise levels (by using different blank scan flux $I_0$).  For training, we use LDCT images with  $I_0=8\times 10^4$.  For testing, we use LDCT images with different noise levels (by using $I_0=5\times 10^4, 6.5\times 10^4, 8\times 10^4$,  and $9.5\times 10^4$). We intentionally carried out training and testing with different noise levels to demonstrate that the proposed method effectively remove noises without causing a deterioration of the underlying morphological information.  Fig. \ref{simu_results_different noise} and Table \ref{table2} show the results of the proposed method for different noise levels. In Table \ref{table2}, the average PSNR, SSIM, and MSE between references and LDCT/corrected images are calculated for the test dataset (40 LDCT images of 2 patients). In terms of MSE, SSIM and PSNR, the proposed method reduces the noise in LDCT images and achieves the best performance at the lowest noise level ($I_0=9.5\times 10^4$). In the case of the highest noise level ($I_0=5\times 10^4$), overall HU values in the corrected images are less than the reference HU values, resulting in the highest MSE and lowest PSNR. On the other hand, the SSIM is  similar at all noise levels.  The robustness of the proposed method can be improved if LDCT images at different noise levels are used for training, as investigated in \cite{Chen2017a}.

\begin{figure*}[ht!]
	\centering
	\includegraphics[width=0.9\textwidth]{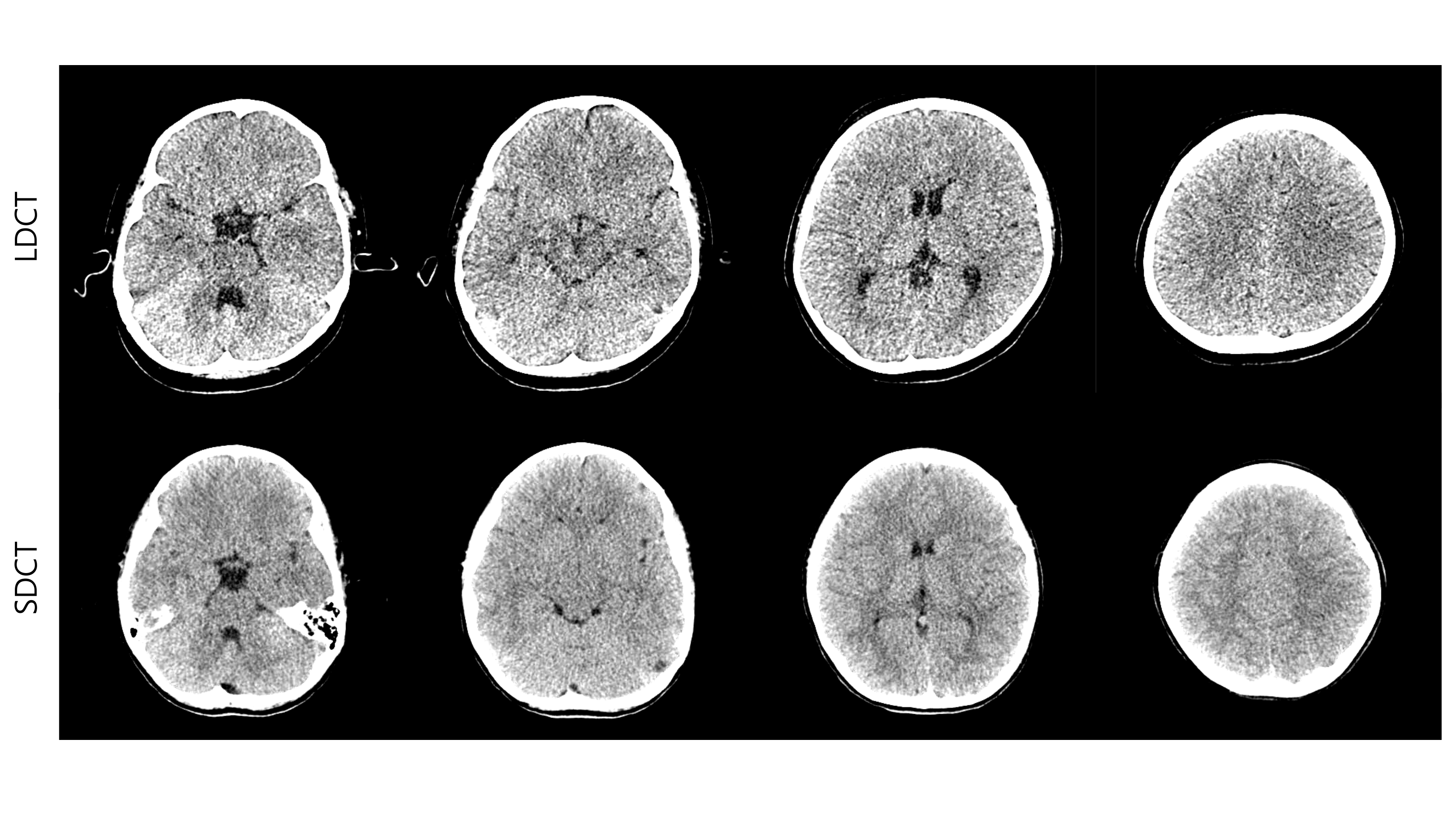}
	\caption{Unpaired training datasets. (top) LDCT images acquired at 100 kVp, 190 reference mAs; (bottom) SDCT images acquired at 120 kVp and 200 mAs  (C=25 HU/ W=50 HU for all CT images)}
	\label{examples_unpair}
\end{figure*}

\begin{figure*}[ht!]
	\includegraphics[width=0.9\textwidth]{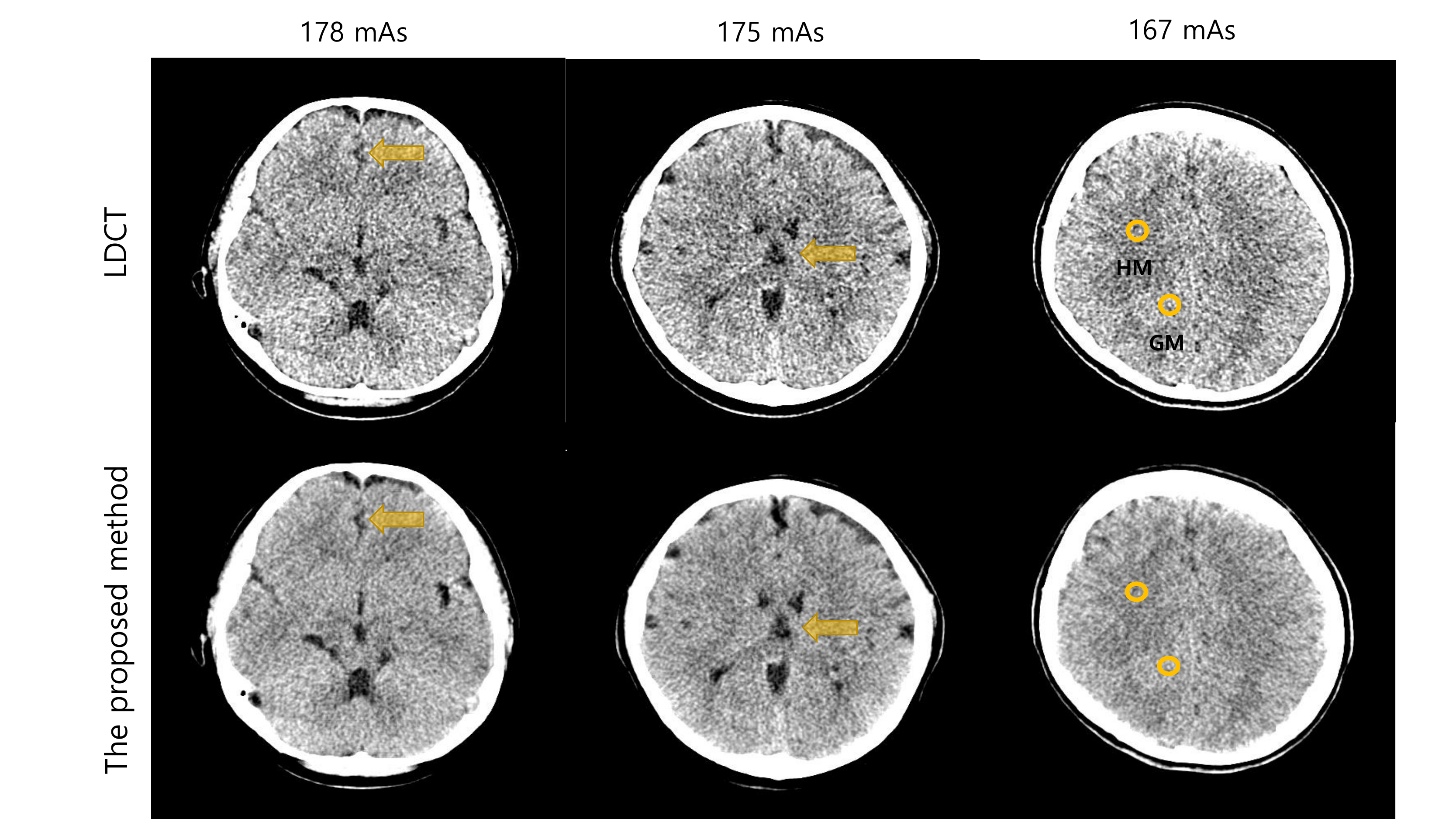}
	\caption{Performance of the proposed method for clinical LDCT images. The first row shows test LDCT images acquired at the different tube current of 178, 175, and 167 mAs, respectively (from left to right). The second row shows the results of the proposed method. On the marked ROIs in yellow, CNR and SNR are calculated in Table \ref{table3}. (C=25 HU/ W=50 HU for all CT images)}
	\label{clinical_results}
\end{figure*}

Fig.   \ref{clinical_results} shows the performance of the clinical experiment for brain CT images. The first row shows the brain LDCT images that are acquired at 178, 175, and 167 mAs. Severe noise artifacts occur due to the low radiation dose. The second row shows the denoising results of the proposed method based on unpaired LDCT and SDCT images that are obtained from different patients. Some examples of unpaired brain CT images used for training are illustrated in Fig.   \ref{examples_unpair}.

For quantitative analysis, we compute the SNR and CNR for test images (60 LDCT images of 3 patients) that were not used for training. The results are summarized in Table \ref{table3}. In this study, the mean and SD of CT number are measured at manually selected ROIs on white matter (WM) and gray matter (GM) structures (Fig. \ref{clinical_results}). The size and shape of each ROI are kept constant. Because no reference image is available for each LDCT image, the SNR for each ROI is calculated as the mean attenuation (in Hounsfield units) divided by the SD \cite{Rapalino2012}.  The CNR between the GM and WM regions is calculated as the difference in the mean attenuation of the two regions divided by the square root of the sum of their variances \cite{Rapalino2012}. A paired t-test with a significance level of $p<0.05$ is performed to assess the statistical significance between the LDCT images and the corrected images. The proposed method increases the SNR value of GM from $4.13\pm 0.48$ to $6.44\pm 0.69~(p<0.001)$, SNR value of HM from $3.49\pm 0.41$ to $5.53\pm 0.59~(p<0.001)$, and CNR value from $0.56\pm 0.19$ to $0.78\pm 0.23~(p<0.001)$. These results demonstrate that the proposed method reduces quantum noise in the corrected images. Even for general noise artifacts (not modeled by Gaussian noise), the proposed method significantly reduces the noise in terms of SNR and CNR.  

\section{Discussion and Conclusion}\label{Discussions}

This study proposes an unpaired deep-learning method of image denoising for LDCT. The method approximates MAP estimation using a GAN framework, and can incorporate prior information on target SDCT images from a data distribution. Under the assumption of Gaussian noise, the $\ell_2$ constraint is added to the GAN framework. Numerical simulations show that the proposed deep learning method with unpaired datasets performs at least compatibly to supervised methods using  paired datasets, in terms of PSNR and MSE (see Fig.   \ref{simu_images}). Clinical results also show that the proposed method enhances LDCT image quality even with unpaired SDCT and LDCT images (see Fig.   \ref{clinical_results}). This type of approach was first used  in amortized MAP inference \cite{Sonderby2017} to solve the super-resolution problem. Note that the proposed objective function \eref{our_model1} is different from that of \cite{Sonderby2017} as  it is derived from the definition of $f$-GAN \cite{Nowozin2016}.

The proposed approach retains some issues requiring further research. One such issue is illustrated by the denoising effect with respect to the choice of samples from the data distribution in Figs. \ref{map-prior} and \ref{efficiency}. This shows that, depending on the samples used for training, the proposed method can remove details,  produce a plausible fake, or simply collapse when generating the denoised images. Therefore, it is essential for the image denoising technique to choose effective training datasets to reflect the appropriate image priors.

Note that the additive Gaussian noise model in \eref{map_gaussian} is not accurate enough.  We adopted this model to simplify the mathematical analysis and to facilitate the derivation of the proposed method. Nevertheless, according to our experiments, the proposed method can learn the target SDCT image features from the data distribution, thus preserving the morphologically important information of the LDCT image and alleviating quantum noise, as well as Gaussian noise, as shown in Fig. \ref{simu_results_quantum noise} and \ref{clinical_results}. The proposed approach can be improved by applying a denoising algorithm that incorporates an accurate noise model of LDCT; for example, other constraints such as perceptual loss \cite{Yang2018} can be applied to the GAN framework in \eref{model-practice} to learn high-level features. In addition, an effective computational method is required for clinical applications. Given that the proposed method works a patch-by-patch, it is more time-consuming than conventional methods that work on an entire image at a time; it takes a few seconds to obtain a single $512\times 512$ denoised CT image. This  can be reduced by using, for example, parallel computation.

Other than the issues mentioned above, future work will also focus on broadening the scope of the proposed method to provide a solution to the other CT reconstruction problems. Such problems include a reduction of the metal artifacts and scattering. This would grant the proposed method the potential to resolve the fundamental difficulty in applying deep learning approaches in X-ray CT (i.e., collecting paired CT image datasets). In addition, it would be  interesting to focus on the quantitative analysis of the relationship between the estimated image and the training datasets.

\section{Appendix}
\subsection{Derivation of MAP for additive Gaussian noise} \label{map-derivation}
We follow the assumption of Section \eref{Method}. That is, the noisy image $z$ is
decomposed into a desired denoised image $x^* \sim p_x$ and additive Gaussian noise $\eps$ of variance $\sigma$ at each pixel.
For the MAP approach, we want to estimate $x^*$ by maximizing the conditional probability $p_{x^*|z}(y|z)$ with respect to the estimation $y$. Bayes rule provides that
\begin{align} \label{map}
	\arg\max_y \log p_{x^*|z}(y|z) = \arg\max_y \left[ \log p_{x}(y)+\log p_{z|x^*}(z|y) \right].
\end{align}
Note that $p_x(y) = p_{x^*}(y)$.
Since $\eps$ is independent of the original image $x^*$, we have
\begin{align}\label{gaussian}
  p_{z|x^*}(z|y) = p_{\eps|x^*} (z - y|y) = p_{\eps}(z-y) = \frac{1}{\sqrt{2\pi\sigma^2}}e^{-\f{\|z-y\|_2^2}{2\sigma^2}}.
\end{align}
It follows from \eref{map} and \eref{gaussian} that the denoised image $y$ is
\begin{align}
\arg\max_y \log p_{x^*|z}(y|z)& = \arg\max_y \left[\log p_{x}(y) -\lambda \|y-z\|_2^2\right]
\end{align}
where $\lambda=\f{1}{2\sigma^2}$.


\bibliographystyle{IEEEtran}

\end{document}